\documentclass[10pt,twocolumn,letterpaper]{article}

\usepackage{cvpr}
\usepackage{times}
\usepackage{epsfig}
\usepackage{graphicx}
\usepackage{amsmath}
\usepackage{amssymb}

\usepackage[pagebackref=true,breaklinks=true,letterpaper=true,colorlinks,bookmarks=false]{hyperref}

\cvprfinalcopy 

\def\cvprPaperID{3995} 

\ifcvprfinal\fi
\begin{document}

\title{Adversarial Feature Augmentation for Unsupervised Domain Adaptation}

\author{Riccardo Volpi\textsuperscript{1}, Pietro Morerio\textsuperscript{1}, Silvio Savarese\textsuperscript{2}, Vittorio Murino\textsuperscript{1,3}\\
{\tt\small \{riccardo.volpi,pietro.morerio,vittorio.murino\}@iit.it, ssilvio@stanford.edu}\\~\\
\textsuperscript{1}Pattern Analysis \& Computer Vision - Istituto Italiano di Tecnologia\\
\textsuperscript{2}Stanford Vision and Learning Lab - Stanford University\\
\textsuperscript{3}Computer Science Department - Universit\`a di Verona\\
}

\maketitle

\begin{abstract}
Recent works showed that Generative Adversarial Networks (GANs) can be successfully applied in unsupervised domain adaptation, where, given a labeled source dataset and an unlabeled target dataset, the goal is to train powerful classifiers for the target samples. In particular, it was shown that a GAN objective function can be used to learn target features indistinguishable from the source ones. In this work, we extend this framework by (i) forcing the learned feature extractor to be domain-invariant, and (ii) training it through data augmentation in the feature space, namely performing feature augmentation. While data augmentation in the image space is a well established technique in deep learning, feature augmentation has not yet received the same level of attention. We accomplish it by means of a feature generator trained by playing the GAN minimax game against source features. Results show that both enforcing domain-invariance and performing feature augmentation lead to superior or comparable performance to state-of-the-art results in several unsupervised domain adaptation benchmarks.\\~\\
\end{abstract}

\section{Introduction}\label{sec:intro}

Generative adversarial networks (GANs \cite{GAN}) are models capable of mapping noise vectors into realistic samples from a data distribution. GANs are defined by two neural networks, a generator and a discriminator, and the training procedure is a minimax game where the generator is optimized to fool the discriminator, and the discriminator is optimized to correctly classify generated samples from actual training samples. Recently, this framework proved to be able to generate images with impressive accuracy \cite{DCGAN}, to generate videos from static frames \cite{DynGAN}, and to translate images from one style to another \cite{DTN,CoGAN,GOOGLE,UITITN}.

Furthermore, GANs have been exploited in the context of unsupervised domain adaptation. Here, a source (labeled) dataset and a target (unlabeled) dataset are considered, which are separated by the so-called domain shift \cite{NameTheDataset}, \ie, they are drawn from two different data distributions. Unsupervised domain adaptation aims at building models that are able to correctly classify target samples, despite the domain shift. In this framework, adversarial training has been used (i) to learn feature extractors that map target samples in a feature space indistinguishable from the one where source samples are mapped \cite{Ganin,ADDA}, and (ii) to develop image-to-image translation algorithms \cite{DTN,CoGAN,GOOGLE,UITITN} aimed at converting source images in a style that resembles that of the target image domain. 

In this paper, we build on the work by Tzeng et al. \cite{ADDA}, which proposes to use a GAN objective to learn target features that are indistinguishable from the source ones, leading to a pair of feature extractors, one for the source and one for the target samples. We extend this approach in two directions: (a) we force domain-invariance in a \emph{single} feature extractor trained through GANs, and (b) we perform data augmentation in the feature space (\ie, \textit{feature augmentation}), by defining a more complex minimax game. 
More specifically, we perform feature augmentation by devising a feature generator trained with a Conditional GAN (CGAN \cite{CGAN}). The minimax game is here played with features instead of images, allowing to generate features conditioned to the desired classes. The CGAN generator is thus able to learn the class distribution in the feature space, and therefore to generate an arbitrary number of labeled feature vectors.
Our results show that forcing domain-invariance and augmenting features are both valuable approaches in the unsupervised domain adaptation setting, leading to higher classification accuracies.
\\~\\
In summary, the \textit{main contributions} of this paper are the following:
\begin{enumerate}
\item Introducing for the first time the use of GANs to perform data augmentation in the feature space.
\item Proposing a new method for unsupervised domain adaptation, based on feature augmentation and (source/target) feature domain-invariance.
\item Evaluating the proposed method on unsupervised domain adaptation benchmarks (cross-dataset digit classification and cross-modal object classification), obtaining results which are superior or comparable to current state-of-the-art in most of the addressed tasks. 
\end{enumerate}
The remaining of the paper is organized as follows. Section~\ref{sec:rel} is dedicated to the related work. The models and the training procedure are presented in Section~\ref{sec:model}. In Section~\ref{sec:datasets}, the datasets used for the analysis and method's validation are described. The experiments and associated results are detailed in  Section~\ref{sec:exp}. Finally, conclusive remarks are drawn in Section~\ref{sec:conclusion}.

\section{Related work}\label{sec:rel}
The work related to our proposed method is focused on GAN research and on modern domain adaptation techniques (\ie, based on deep learning).\\~\\
\textbf{Generative adversarial networks.} In the original formulation by Goodfellow et al. \cite{GAN}, a GAN model is trained through a minimax game between a generator, that maps noise vectors in the image space, and a discriminator, trained to discriminate generated images from real ones. Several other papers address ways to control what GANs generate \cite{CGAN,InfoGAN,WhatWhereGAN}. In particular, CGANs \cite{CGAN} allow to condition on the desired classes, from which samples are generated. Other works \cite{ALI,BiDir} propose to learn inference  by playing a minimax game against features. In these works, trained models are feature extractors that map images into the feature space, not feature generators, which is \textit{our primary goal}.

Performing feature augmentation through GANs is one of the original aspects of our approach. We propose a generator able to generate features from noise vectors and label codes, via a CGAN \cite{CGAN} framework, playing a minimax game with features extracted from a pre-trained model instead of images.
\\~\\    
\textbf{Unsupervised domain adaptation.} Ganin and Lempitsky \cite{Ganin} propose a neural network (Domain-Adversarial Neural Network, DANN) where a ConvNet-based \cite{MNIST} feature extractor is optimized to both correctly classify source samples and have domain-invariant features, through adversarial training. Different works \cite{MMD1,MMD2} aim at minimizing the Maximum Mean Discrepancy \cite{MMD} between features extracted from source and target samples, training a classifier to correctly classify source samples while minimizing this measure. Bousmalis et al. \cite{DSN} propose to learn image representations divided in two components, one shared across domains and one private, following the hypothesis that modeling unique elements in each domain can help to extract features which are domain-invariant. Tzeng et al. \cite{ADDA} use GANs to train an encoder for target samples, by making the features extracted with this model indistinguishable from the ones extracted through an encoder trained with source samples. The last layer of the latter can then be used for both encoders to infer labels. Saito et al. \cite{Tri} propose an asymmetric tri-training where pseudo-labels are inferred and exploited for target samples during training. In particular, two networks are trained to assign labels to target samples and one to obtain target-discriminative features. Haeusser et al. \cite{ADA} propose to exploit associations between source and target features during training, to maximize the domain-invariance of the learned features while minimizing the error on source samples. 
\begin{figure*}[h]
	\begin{center}
		\includegraphics[width=\textwidth]{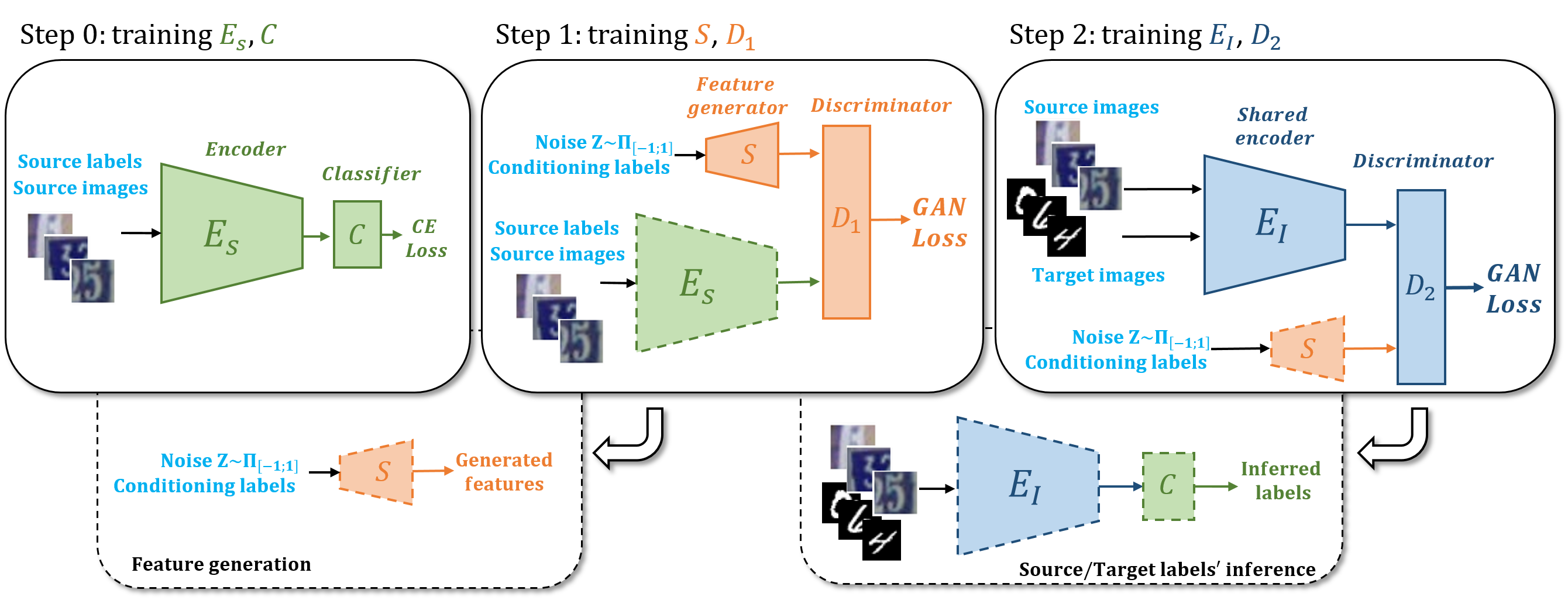}
	\end{center}
	\caption{Training procedure, representing the steps described in Section~\ref{sec:training}. Solid lines indicate that the module is being trained, dashed lines indicate that the module is already trained (from previous steps). All modules are neural networks, whose architectures are detailed in Section~\ref{sec:arch}. Smaller, dashed panels in the bottom indicate how to generate features (\textit{left}) and how to infer source or target labels (\textit{right}).} 

\label{model}
\end{figure*}

Recently, several image-to-image translation methods have been proposed to solve unsupervised domain adaptation tasks. Taigman et al. \cite{DTN} propose the Domain Transfer Network (DTN), that allows to translate images from a source domain to a target one, under a $f$-constancy constraint, where $f$ is a generic function that maps images in a feature space. Translated images result portrayed in the target images' style, while maintaining the content of the images fed in input.  Liu and Tuzel \cite{CoGAN} introduce Coupled GAN (CoGAN), an extension of GAN that allows to model a joint distribution $P(X,Y)$ and to generate couples of images from noise vectors, one belonging to $P(X)$ and one to $P(Y)$. This model can be applied to image-to-image translation tasks: fixing one image, the noise vector that most likely could have generated that picture can be inferred and, feeding it to the model, the second image is generated. Bousmalis et al. \cite{GOOGLE} propose to train an image-to-image translation network relying on both a GAN loss and a \textit{task-specific} loss (and in problems with prior knowledge, also a \textit{content-specific} loss). The resulting network takes in input both an image and a noise vector, that allows to generate a potentially infinite number of target images. Liu et al. \cite{UITITN} propose UNIT, an extension of CoGAN that relies on both GANs and Variational Auto-Encoders, and makes the assumption of a shared latent space. Image-to-image translation methods \cite{DTN,CoGAN,GOOGLE,UITITN} are applied to unsupervised domain adaptation by generating target images and training classifiers directly on them.

The domain-invariant feature extractor we designed is inspired by Tzeng et al. \cite{ADDA}, with two main differences. First, we play the minimax game against features which are \textit{generated} by a pre-trained model, thus performing \textit{feature augmentation}. Second, we train the feature extractor in order to make it work for both source and target samples (thus achieving \textit{domain-invariance}), avoiding catastrophic forgetting. Both modifications lead to higher accuracies in classifying target samples, as we will show in Section~\ref{sec:exp}. Domain-invariance also allows to use the same feature extractor for both source and target samples, while in Tzeng et al. \cite{ADDA} two different encoders are required. 

\section{Model}\label{sec:model}
Our goal is to train a domain-invariant feature extractor ($E_I$), whose training procedure is made more robust by data augmentation in the space of source features. The training procedure we designed to accomplish our intent is based on three different steps, depicted in Figure~\ref{model}. First, we need to train a feature extractor on source data ($C \circ E_s$). This step is necessary because we need a reference feature space  and a reference classifier that performs well on it. Secondly, we need to train a feature generator ($S$) to perform data augmentation in the source feature space. We can train it by playing a GAN minimax game against features extracted through $E_S$. Finally, we can train a domain-invariant feature extractor ($E_I$) by playing a GAN minimax game against features generated through $S$. This module can then be combined with the softmax layer previously trained ($C \circ  E_I$) to perform inference on both source and target samples. All modules are neural networks trained by backpropagation \cite{BackProp}. In the following sections, we detail how each Step is performed, how new features can be generated, and how source/target labels can be inferred.

\subsection{Training}\label{sec:training}
\textbf{Step 0.} The model $C \circ E_s$ is trained to classify source samples. $E_s$ represents a ConvNet feature extractor and $C$ represents a fully connected  softmax layer, with a size that depends on the problem. The optimization problem consists in the minimization of the following cross-entropy loss (\textit{CE Loss} in Figure~\ref{model}):
\begin{equation}
\min_{\theta_{E_s}, \theta_C} \ell_{0} = \mathbb{E}_{(x_i,y_i) \sim (X_s,Y_s)} H(C \circ E_s(x_i),y_i),
\end{equation}
where $\theta_{E_s}$ and $\theta_C$ indicate the parameters of $E_s$ and $C$, respectively, $X_s$, $Y_s$ are the distributions of source samples ($x_i$) and source labels ($y_i$), respectively, and $H$ represents the softmax cross-entropy function.\\~\\
\textbf{Step 1.} The model $S$ is trained to generate feature samples that resemble the source features. Exploiting the CGAN framework, the following minimax game is defined:
\begin{align}
\min_{\theta_S} \max_{\theta_{D_1}} \ell_{1} &= \mathbb{E}_{(z,y_i) \sim (p_z(z),Y_s)} \Vert D_1(S(z \vert\vert y_i) \vert\vert y_i) - 1 \Vert^2 \nonumber \\
& + \mathbb{E}_{(x_i,y_i) \sim (X_s,Y_s)} \Vert\vert D_1(E_s(x_i) \vert\vert y_i) \Vert^2, 
\end{align}
where $\theta_S$ and $\theta_{D_1}$ indicate the parameters of $S$ and $D_1$, respectively, $p_z(z)$ is the distribution\footnote{Uniform in the  range $[-1,1]$ throughout this work.} from which noise samples are drawn, and $\Vert$ denotes a concatenation operation. In this and the following steps, we relied on Least Squares GANs \cite{LSGAN} since we observed more stability during training.
\\~\\
\textbf{Feature generation.} In order to generate an arbitrary number of new feature samples, we only need $S$, which takes as input the concatenation of a noise vector and a \textit{one-hot} label code, and outputs a feature vector from the desired class:  
\begin{equation}\label{eq:feat}
\vspace{-10pt}
F(z\vert y) = S(z \vert\vert y)  
\end{equation}\\
where $z \sim p_z(z)$ and $F$ is a feature vector belonging to the class label associated with $y$ (dashed box in Figure~\ref{model}, \textit{left}).\\~\\
\textbf{Step 2.} The domain-invariant encoder $E_I$ is trained via the following minimax game, after being initialized with weights optimized on Step 0 (note that $E_S$ and $E_I$ have the same architecture), a requirement to reach optimal convergence:
\begin{align}
\min_{\theta_{E_I}} \max_{\theta_{D_2}} \ell_{2} &= \mathbb{E}_{x_i \sim X_s \cup X_t} \Vert D_2(E_I(x_i)) - 1 \Vert^2\\ 
&\quad+ \mathbb{E}_{(z,y_i) \sim (p_z(z),Y_s)} \Vert D_2(S(z \vert\vert y_i)) \Vert^2, \nonumber
\end{align}
where $\theta_{E_I}$ and $\theta_{D_2}$ indicate the parameters of $E_I$ and $D_2$, respectively. Since the model $E_I$ is trained using both source and target domains, the feature extractor results domain-invariant. In particular, it maps both source and target samples in a common feature space, where features are indistinguishable from the ones generated through $S$. Being the latter trained to produce features indistinguishable from the source ones, the feature extractor $E_I$ can be combined with the classification layer of Step 0 ($C$) and used for inference (as in Tzeng et al. \cite{ADDA}):

\begin{equation}\label{eq:inference}
\tilde y_i = C \circ E_I(x_i),
\end{equation}
where $x_i$ is a generic image from the source or the target data distribution and $\tilde y_i$ is the inferred label (dashed box in Figure~\ref{model}, \textit{right}).

\section{Datasets}\label{sec:datasets}

\begin{figure*}[t]
	\begin{center}
		\includegraphics[width=0.9\textwidth]{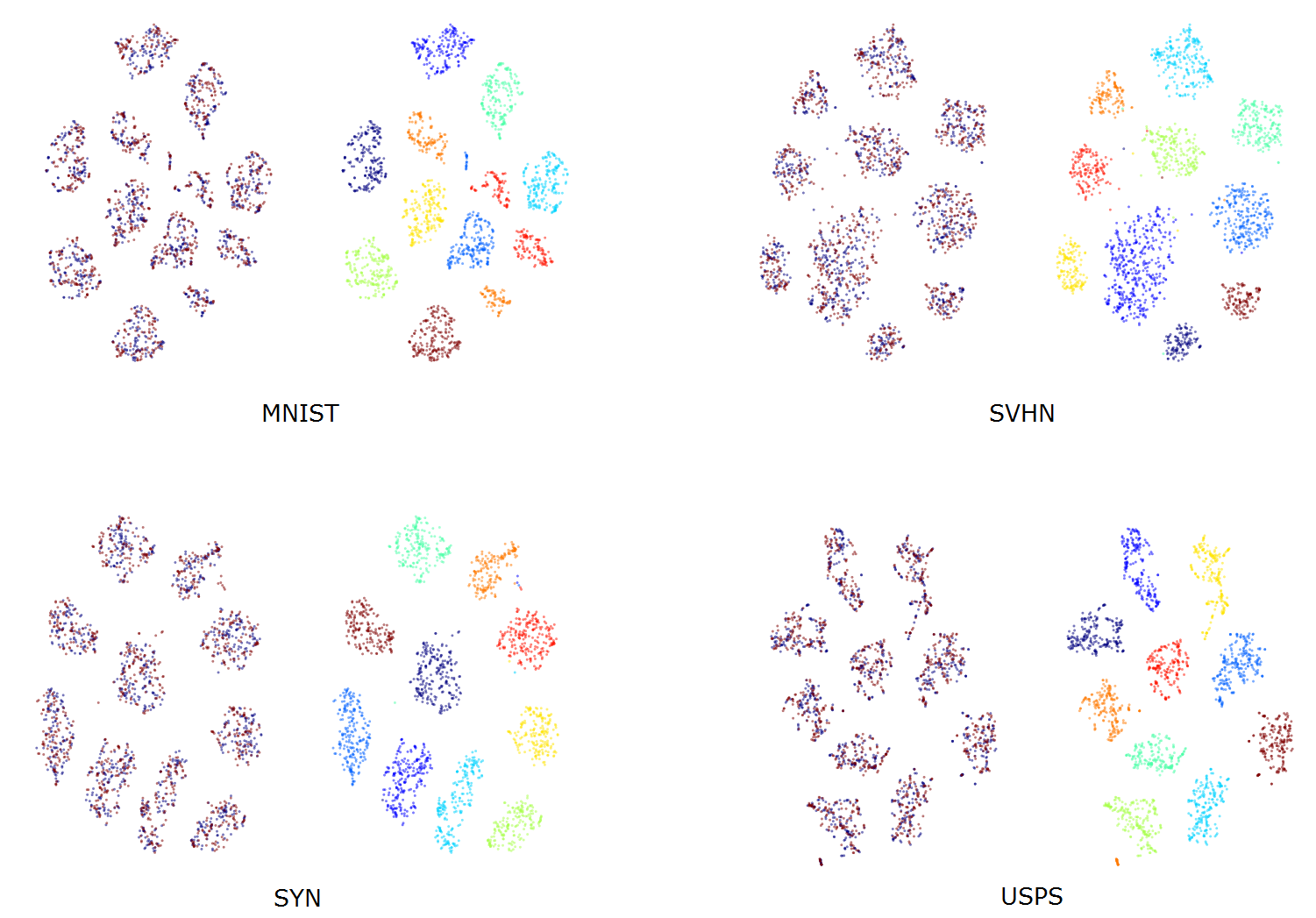}
	\end{center}
	\caption{t-SNE plots of features associated with different adopted datasets (MNIST, SVHN, SYN, USPS). For each dataset, in the \textit{left} part of the panels, red and blue dots indicate real and generated features, respectively. In the \textit{right} part of the panels, different colors indicate different classes.} 
	\label{tsne}
\end{figure*} 

To evaluate our approach, we used several benchmark splits of public source/target datasets adopted in domain adaptation.\\~\\ 
\textbf{MNIST $\leftrightarrow$ USPS.} Both datasets consist of white digits on a solid black background. We tested two different protocols: the first one (P1)  consists in sampling $2,000$ MNIST \cite{MNIST} images and $1,800$ USPS \cite{USPS} images. The second one (P2) consists in using the whole MNIST training set, $50,000$ images, and dividing USPS in $6,562$ images for training, $2.007$ for testing, and $729$ for validation. For P1, we tested the two directions of the split (MNIST $\rightarrow$ USPS and MNIST $\leftarrow$ USPS). For P2, we tested only MNIST $\rightarrow$ USPS, and we avoided to use the validation set in this case, too. In both experimental protocols, we resized USPS digits to $28 \times 28$ pixels, which is the MNIST images' size.
\\~\\
\textbf{SVHN $\rightarrow$ MNIST.} SVHN \cite{SVHN} is built with real images of Street View House Numbers. We used the whole training sets of both datasets, following the standard protocol for unsupervised domain adaptation (SVHN training set contains $73,257$ images), and tested on MNIST test set. We resized MNIST images to $32 \times 32$ pixels and converted SVHN to grayscale. We did not use the extra set of SVHN. 
\\~\\
\textbf{SYN DIGITS $\rightarrow$ SVHN.} This split represents a synthetic-to-real domain adaptation problem, of great interest for research in computer vision since, quite often, generating labeled synthetic data requires less effort than obtaining large labeled dataset with real examples. SYN DIGITS \cite{Ganin} contains $500,000$ images belonging to the same SVHN classes. We tested on SVHN test set.
\\~\\  
\textbf{NYUD (RGB $\rightarrow$ D).} This modality adaptation problem was proposed by Tzeng et al. \cite{ADDA}. The dataset is gathered by cropping out tight bounding boxes around instances of 19 object classes present in the NYUD \cite{NYUD} dataset. It comprises 2,186 labeled source (RGB) images and 2,401 unlabeled target (HHA-encoded \cite{HHA}) depth images. Note that these are obtained from two different splits of the original dataset, to ensure that the same instance is not seen in both domains. The adaptation task is extremely challenging, due to the very different domains, the limited number of examples (especially for some classes), and the low resolution of the cropped bounding boxes.

\section{Experiments}\label{sec:exp}

In this section, we evaluate our approach. First, we show that our model $S$ is able to generate consistent and discriminant feature vectors conditioned on the desired classes. Second, we report an ablation study 
to figure out the benefits brought by the different steps that compose our approach. Finally, we compare our method with competing algorithms on unsupervised domain adaptation tasks.

\subsection{Architectures}\label{sec:arch}

A detailed description of architectures and hyperparameters used (learning rate, batch sizes, \textit{etc.}) is reported in the Supplementary Material. We provide here the details necessary for a basic understanding of the experiments.\protect\footnote{Models were implemented using Tensorflow, and training procedures were performed on a NVIDIA Titan X GPU. Code: \url{https://github.com/ricvolpi/adversarial-feature-augmentation}}.
 
$S$ is built by the repetition of two blocks, each defined by a fully connected layer, a Batch Normalization layer \cite{BatchNorm}, and a Dropout layer \cite{Dropout}, followed by a fully connected layer with \textit{tanh} activation functions. $D_1$ is a one-hidden-layer neural network, with a \textit{sigmoid} hidden unit as output layer. We defined $E_S$ and $E_I$ following standard architectures used in unsupervised domain adaptation \cite{Ganin}. In particular, for SVHN $\rightarrow$ MNIST, MNIST $\rightarrow$ USPS and USPS $\rightarrow$ MNIST, we defined the network as conv-pool-conv-pool-fc-fc-softmax (with Dropout \cite{Dropout} on fully connected layers for MNIST $\leftrightarrow$ USPS experiments). For SYN $\rightarrow$ SVHN, conv-pool-conv-pool-conv-fc-fc-softmax. For the NYUD experiment, in order to be comparable with \cite{ADDA}, we  used a VGG-16 \cite{VGG16} pretrained on ImageNet \cite{ImageNet}. The final feature dimensionality (\eg, the size of the feature vector fed to the softmax layer) was set to $128$ for all experiments, except for SYN $\rightarrow$ SVHN ($256$). $D_2$ is built with two or three fully connected layers (depending on the experiment) with a \textit{sigmoid} unit on top. Note that for the NYUD experiment we used three hidden layers, while Tzeng et al. \cite{ADDA} built the discriminator with two, since our method requires an additional one to reach convergence. For all our experiments, we used Adam optimizer \cite{Adam} with momentum set to $0.95$. \textit{ReLU} \cite{ReLU} units were used throughout the architectures, except for last layers of discriminators, defined as \textit{sigmoid} units, last layer of $S$, whose activation functions are \textit{tanh}, and $D_2$, which was built with \textit{Leaky ReLU} units, in agreement with the findings of Radford et al. \cite{DCGAN}.

\begin{table}[hb!]
	\centering
	\caption{\textit{Second} column: number of activation patterns (APs) among the features extracted from training data. \textit{Third} column: number of APs that $S$ is able to generate. \textit{Fourth} column: classification accuracy of the generated features, accordingly to given labels.}
    \vspace{5pt}
	\label{tab:feat}
	\begin{tabular}{|r||l|c|l|}
		\hline
		Dataset & \#APs $E_S(x)$ & \#APs $S(z \Vert y)$                     & Accuracy \\ \hline\hline
		SVHN    & $69,625$     &                                  & $0.974$    \\ 
		USPS    & $1,422$      &     $\mathbf{\sim 10^6}$                       & $0.998$    \\ 
		MNIST   & $1,910$      &                                  & $0.995$    \\ \cline{3-3}
        NYUD    & $19$         &     $\mathbf{\sim 10^3}$         & $0.998$   \\ \hline
	\end{tabular}
\end{table}

\subsection{Generating features}\label{subsec:feat}

\begin{figure*}[t]
	\begin{center}
		\includegraphics[width=\textwidth]{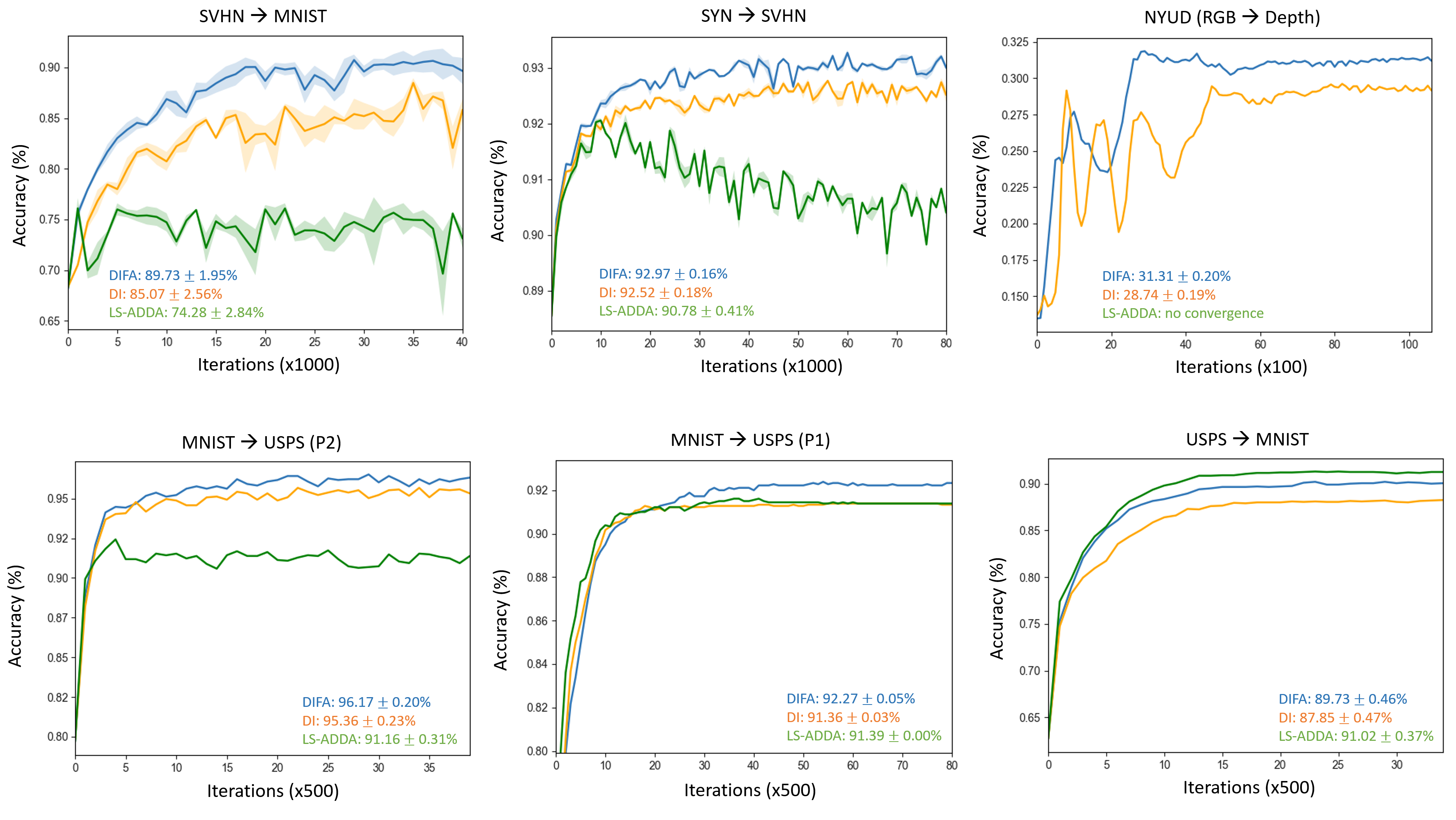}
	\end{center}
	\caption{Accuracies on target samples evaluated throughout the training of the feature extractors of \textit{LS-ADDA} (\textit{green}), \textit{DI} (\textit{orange}) and DIFA (\textit{blue}). Inference was performed by combining the feature extractor being learned with $C$ of Step 0, Section~\ref{sec:training}. In the NYUD experiment the green curve is missing due to non-convergence of \textit{LS-ADDA}. SVHN $\rightarrow$ MNIST and SYN $\rightarrow$ SVHN plots were obtained averaging over three different runs; confidence bands are portrayed.} 
	\label{abl}
\end{figure*}

We qualitatively show with t-SNE \cite{TSNE} that we can generate feature vectors from the desired classes, after having trained $S$ as described in Section~\ref{sec:training}. Figure~\ref{tsne} shows comparisons between real and generated features for different datasets. For each dataset, two identical point clouds are represented: the \textit{bi-color} side (at the left of each panel), highlights real and generated samples (in red and blue, respectively); the \textit{multi-color} side (at the right of each panel) highlights instead the different classes. From a qualitative point of view, real and generate features appear indistinguishable, and class structure is preserved. To quantitatively measure the quality of the features generated, we fed them to the classifier $C$ trained with the original samples for class estimation. Table~\ref{tab:feat} (fourth column) shows that such features are also quantitatively reliable, and this is valid for all the datasets considered.\\~\\ 
\textbf{Feature augmentation.} Finally, we are interested in evaluating the variability of the features generated through $S$ to figure out whether (i) the model is memorizing the features from the training set, and (ii) it is realistic to assume that we are performing data augmentation in the feature space. To shed light on these two questions, we decided to perform the following empirical test: we counted the number of activation patterns (APs) that $S$ is able to generate, and compared it with the ones intrinsically available in the original dataset. An activation pattern is defined by thresholding the output of the activation functions of the hidden state of a network. Raghu et al. \cite{Express} defined this concept for \textit{ReLUs} \cite{ReLU}, where values greater than zero are set to one, the others to zero. For our purposes, we can apply the same rule even if we are using \textit{tanh} activation functions. For example, SVHN has $73,257$ samples that - with the feature extractor we used for our experiments - correspond to $69,625$ activation patterns. $S$ can instead generate a number of activation patterns in the order of $10^6$ (counted empirically, feeding noise to $S$ till saturation), indistinguishable from the original ones due to the training procedure defined in Section~\ref{sec:training}. Table~\ref{tab:feat} reports the results associated with the other datasets considered. Interestingly, activation patterns associated with the $2,186$ source samples of NYUD are only $19$: each pattern is associated with a different class. This is most likely due to overfitting: the network is already explicitly encoding classes at feature level. However, the generator $S$ can enrich the feature set to a broad extent.

\subsection{Ablation study}
\begin{table*}[ht]
\centering
\small
\caption{Comparison of our method with competing algorithms. The row \textit{LS-ADDA} lists results obtained by our implementation of Least Squares ADDA. The row \textit{Ours (DI)} refers to our approach in which only domain-invariance is imposed. 
The row \textit{Ours (DIFA)} refers to our full proposed method, which includes feature augmentation. 
(*) DTN \cite{DTN} and UNIT \cite{UITITN} use extra SVHN data ($531,131$ images). (**) Protocols P1 and P2 are mixed in the results section of Bousmalis et al. \cite{GOOGLE}. Convergence not reached is indicated as \textit{no conv.}}
   \vspace{5pt}
\label{tab:res}
\begin{tabular}{lllllll}
                         & \footnotesize SVHN$\rightarrow$MNIST                          & \footnotesize MNIST$\rightarrow$USPS$_{P1}$              & \footnotesize MNIST$\rightarrow$USPS$_{P2}$              & \footnotesize USPS$\rightarrow$MNIST                   & \footnotesize SYN$\rightarrow$SVHN                     & \footnotesize NYUD                       \\ \hline
\multicolumn{1}{|l|}{Source}     & \multicolumn{1}{l|}{0.682}          & \multicolumn{1}{l|}{0.723}   & \multicolumn{1}{l|}{0.797}   & \multicolumn{1}{l|}{0.627}   & \multicolumn{1}{l|}{0.885}   & \multicolumn{1}{l|}{0.139} \\ \hline \hline
\multicolumn{1}{|l|}{DANN {\footnotesize \cite{Ganin,Ganin2}}}   & \multicolumn{1}{l|}{0.739}          & \multicolumn{1}{l|}{0.771 $\pm$ 0.018 {\footnotesize \cite{ADDA}}}   & \multicolumn{1}{l|}{-}       & \multicolumn{1}{l|}{0.730 $\pm$ 0.020 {\footnotesize \cite{ADDA}}}   & \multicolumn{1}{l|}{0.911}   & \multicolumn{1}{l|}{-}     \\ 
\multicolumn{1}{|l|}{DDC {\footnotesize \cite{ADDA}}}     & \multicolumn{1}{l|}{0.681 $\pm$ 0.003}          & \multicolumn{1}{l|}{0.791 $\pm$ 0.005}   & \multicolumn{1}{l|}{-}       & \multicolumn{1}{l|}{0.665 $\pm$ 0.033}   & \multicolumn{1}{l|}{-}        & \multicolumn{1}{l|}{-}     \\ 
\multicolumn{1}{|l|}{DSN {\footnotesize \cite{DSN}}}     & \multicolumn{1}{l|}{0.827}          & \multicolumn{1}{l|}{-}       & \multicolumn{1}{l|}{-}       & \multicolumn{1}{l|}{-}       & \multicolumn{1}{l|}{0.912}   & \multicolumn{1}{l|}{-}     \\ 
\multicolumn{1}{|l|}{ADDA {\footnotesize \cite{ADDA}}}    & \multicolumn{1}{l|}{0.760 $\pm$ 0.018}          & \multicolumn{1}{l|}{0.894 $\pm$ 0.002}   & \multicolumn{1}{l|}{-}       & \multicolumn{1}{l|}{\textbf{0.901 $\pm$ 0.008}}   & \multicolumn{1}{l|}{-}        & \multicolumn{1}{l|}{0.211} \\ 
\multicolumn{1}{|l|}{Tri {\footnotesize \cite{Tri}}}  & \multicolumn{1}{l|}{0.862}          & \multicolumn{1}{l|}{-}       & \multicolumn{1}{l|}{-}       & \multicolumn{1}{l|}{-}       & \multicolumn{1}{l|}{\textbf{0.931}}   & \multicolumn{1}{l|}{-}     \\ 
\multicolumn{1}{|l|}{DTN {\footnotesize \cite{DTN}}}     & \multicolumn{1}{l|}{0.844*}          & \multicolumn{1}{l|}{-}       & \multicolumn{1}{l|}{-}       & \multicolumn{1}{l|}{-}       & \multicolumn{1}{l|}{-}       & \multicolumn{1}{l|}{-}     \\ 
\multicolumn{1}{|l|}{PixelDA** {\footnotesize \cite{GOOGLE}}}  & \multicolumn{1}{l|}{-}              & \multicolumn{1}{l|}{-}       & \multicolumn{1}{l|}{0.959}   & \multicolumn{1}{l|}{-}       & \multicolumn{1}{l|}{-}       & \multicolumn{1}{l|}{-}     \\ 
\multicolumn{1}{|l|}{UNIT {\footnotesize \cite{UITITN}}}  & \multicolumn{1}{l|}{\textbf{0.905*}}              & \multicolumn{1}{l|}{-}       & \multicolumn{1}{l|}{0.960}   & \multicolumn{1}{l|}{-}       & \multicolumn{1}{l|}{-}       & \multicolumn{1}{l|}{-}     \\ 
\multicolumn{1}{|l|}{CoGANs {\footnotesize \cite{CoGAN}}} & \multicolumn{1}{l|}{no conv. {\footnotesize \cite{ADDA}}} & \multicolumn{1}{l|}{0.912 $\pm$ 0.008}   & \multicolumn{1}{l|}{0.957 {\footnotesize \cite{UITITN}}}        & \multicolumn{1}{l|}{0.891 $\pm$ 0.008}   & \multicolumn{1}{l|}{-}       & \multicolumn{1}{l|}{-}     \\ \hline \hline
\multicolumn{1}{|l|}{\textit{LS-ADDA}} & \multicolumn{1}{l|}{0.743 $\pm$ 0.028}          & \multicolumn{1}{l|}{0.914 $\pm$ 0.000}   & \multicolumn{1}{l|}{0.912 $\pm$ 0.003}   & \multicolumn{1}{l|}{\textbf{0.910 $\pm$ 0.004}}   & \multicolumn{1}{l|}{0.908 $\pm$ 0.004}       & \multicolumn{1}{l|}{no conv.}     \\ 
\multicolumn{1}{|l|}{\textit{Ours {\footnotesize (DI)}}} & \multicolumn{1}{l|}{0.851 $\pm$ 0.026}          & \multicolumn{1}{l|}{0.914 $\pm$ 0.000}   & \multicolumn{1}{l|}{0.954 $\pm$ 0.002}   & \multicolumn{1}{l|}{0.879 $\pm$ 0.005}   & \multicolumn{1}{l|}{0.925 $\pm$ 0.002}       & \multicolumn{1}{l|}{0.287 $\pm$ 0.002}     \\
\multicolumn{1}{|l|}{\textit{Ours {\footnotesize (DIFA)}}} & \multicolumn{1}{l|}{\textbf{0.897 $\pm$ 0.020}}          & \multicolumn{1}{l|}{\textbf{0.923 $\pm$ 0.001}}   & \multicolumn{1}{l|}{0.962 $\pm$ 0.002}   & \multicolumn{1}{l|}{0.897 $\pm$ 0.005}   & \multicolumn{1}{l|}{\textbf{0.930 $\pm$ 0.002}}   & \multicolumn{1}{l|}{\textbf{0.313 $\pm$ 0.002}} \\ \hline \hline
\multicolumn{1}{|l|}{Target}     & \multicolumn{1}{l|}{0.992}        & \multicolumn{1}{l|}{0.999} & \multicolumn{1}{l|}{0.999} & \multicolumn{1}{l|}{0.975} & \multicolumn{1}{l|}{0.913} & \multicolumn{1}{l|}{0.468 {\footnotesize \cite{ADDA}}} \\ \hline
\end{tabular}
\end{table*}

We carried out an ablation study to evaluate the benefit brought by the introduced modifications to the current way of using GAN objectives in unsupervised domain adaptation. Since the Least Squares GAN \cite{LSGAN} framework is required to solve Step 1 and Step 2 of our method (Section~\ref{sec:training}), we re-designed the ADDA algorithm \cite{ADDA} in this framework as a baseline, and from this point we implemented our peculiar contributions, showing that each one favourably concurs to improve performance. We term it \textit{LS-ADDA}, and it is defined by the following minimax game:
\begin{align}
\min_{\theta_{E_t}} \max_{\theta_{D}} \ell &= \mathbb{E}_{x_i \sim X_t} \Vert D(E_t(x_i)) - 1 \Vert^2\\ 
&\quad+ \mathbb{E}_{x_i \sim X_s} \Vert D(E_s(x_i)) \Vert^2, \nonumber
\end{align}
where $E_s$ is the feature extractor trained on source samples (as the one pre-trained in Step 0, Figure~\ref{model}), and $E_t$ is the encoder for the target samples that is being trained. $D$ is the discriminator, as those described in this work.

The second analysis stage lies in imposing domain-invariance, and this is carried out by solving the following minimax problem:
\begin{align}
\min_{\theta_{E_I}} \max_{\theta_{D}} \ell &= \mathbb{E}_{x_i \sim X_s \cup X_t} \Vert D(E_{I}(x_i)) - 1 \Vert^2\\ 
&\quad+ \mathbb{E}_{x_i \sim X_s} \Vert D(E_s(x_i)) \Vert^2, \nonumber
\end{align}
where $E_{I}$ is the \emph{shared} encoder for the source and target samples that is being trained, and the rest of the modules are the same described above. This represents our first notable contribution, which we call \textit{DI} (short for \textit{DI LS-ADDA}, as this architecture introduces domain-invariance to \textit{LS-ADDA}). Finally, the third analysis stage is constituted by our complete proposed approach, in which the minimax game also embeds the feature augmentation procedure (described in Step 2 of Section~\ref{sec:training}). We term it \textit{DIFA} (\textit{Domain-Invariance + Feature Augmentation}). For each of the three architectures proposed in this ablation study, we finally end up with an encoder that can be combined with the module $C$ trained in Step 0 (see Figure~\ref{model}). We tested these algorithms on the benchmark splits detailed in Section~\ref{sec:datasets}. Figure~\ref{abl} shows the evolution of the performance of these three frameworks throughout the minimax games: \textit{green} curves are associated with \textit{LS-ADDA}, \textit{orange} curves are associated with \textit{DI}, and \textit{blue} curves are associated with \textit{DIFA}. 
\begin{table}[]
\centering
\caption{Difference in accuracy between training and test \textit{source} data, by classifying with $C \circ E_S$ and $C \circ E_I$. Source test data is not provided for NYUD \cite{ADDA}. $E_I$ does not experience catastrophic forgetting and generalizes well on unseen source data (test).}
\label{tab:res_di}
\begin{tabular}{|l||c|c|}
\hline
Dataset & $E_S \rightarrow E_I (training)$               & $E_S \rightarrow E_I (test)$                      \\ \hline \hline
USPS    & 0.975 $\rightarrow$ 0.973 & 0.980 $\rightarrow$ 0.979 \\ 
MNIST \tiny{(P1)}   & 1.000 $\rightarrow$ 0.997 & 0.960 $\rightarrow$ 0.961 \\ 
MNIST \tiny{(P2)}   & 0.997 $\rightarrow$ 0.986 & 0.992 $\rightarrow$ 0.984 \\ 
SVHN    & 0.982 $\rightarrow$ 0.883 & 0.905 $\rightarrow$ 0.856 \\ 
SYN     & 0.998 $\rightarrow$ 0.996 & 0.995 $\rightarrow$ 0.994 \\ 
NYUD    & 1.000 $\rightarrow$ 1.000 & \textit{test set n.a.} \\ \hline
\end{tabular}
\end{table}
The values reported in the bottom part of the plots indicate the average and the standard deviation calculated over the final stages of training, \ie, when the minimax game reaches a stability point, despite oscillations. For the splits SVHN $\rightarrow$ MNIST and SYN $\rightarrow$ SVHN, we averaged over three different runs, due to some instability in the equilibriums reached, that can be observed in Figure~\ref{abl}. The general trend is that enforcing domain-invariance (\textit{DI}) brings a first improvement (except in the MNIST $\rightarrow$ USPS (P1) experiment), and feature augmentation (\textit{DIFA}) adds a further increment. In NYUD, \textit{LS-ADDA} cannot converge.

The only exception is USPS $\rightarrow$ MNIST, where \textit{LS-ADDA} is the best performing method. Note that we did not report experiments related to embedding feature augmentation without domain-invariance because it performs poorly, due to high instability.

\subsection{Comparisons with other methods}

Table~\ref{tab:res} reports our findings and results obtained by the other works in the literature. The first row reports accuracies on target data achieved with non-adapted classifiers trained on source data, and the last row reports accuracies on target data achieved with classifiers trained on target data (\textit{oracle}). Our main contributions lie in forcing the domain-invariance in the GAN minimax game (\textit{DI}) and further improving it with feature augmentation (\textit{DIFA}). A difficulty in unsupervised domain adaptation is determine the fair accuracy reached by each method, since cross-validation is not feasible (target labels should be used only to evaluate the method at the end of the training procedure). We believe that a fair way is the one we proposed in the previous section ($mean \pm std$ calculated over the last iterations), since choosing a single value would be arbitrary and unfair in stochastic training procedures (\eg, see SVHN $\rightarrow$ MNIST and SYN $\rightarrow$ SVHN in Figure~\ref{abl}).

Results show that our approach based on domain-invariance and feature augmentation leads to accuracies comparable or higher to current state-of-the-art in several unsupervised domain adaptation benchmarks. Among the splits we tested, the only exception is USPS $\rightarrow$ MNIST, where ADDA \cite{ADDA} and our implementation of it (\textit{LS-ADDA}) perform better - with the drawback of having two different feature extractors for source/target samples. In SVHN $\rightarrow$ MNIST, our approach gives results comparable to current state-of-the-art (UNIT \cite{UITITN}), but it must be noted that the latter was achieved by making use of extra SVHN set ($531,131$ images), making the result difficult to interpret. In MNIST $\rightarrow$ USPS (P2) we perform better or comparably to any other method that was tested on it. Also note that all those methods \cite{GOOGLE,CoGAN,UITITN} rely on the generation of target images to perform adaptation, and that \cite{GOOGLE,UITITN} rely on additional hyperparameters - a severe drawback in unsupervised domain adaptation, where cross-validation is not applicable. In SYN $\rightarrow$ SVHN, our method is statistically comparable with the one proposed by Saito et al. \cite{Tri}. In this case, it is also worth noting that the adapted feature extractor performs better than a neural network trained on SVHN (target) training set (see Table~\ref{tab:res}, last row). This opens a wide range of possibility of using synthetic data, which are much easier to obtain than labeled, real data in real-world applications. In NYUD (RGB $\rightarrow$ Depth), we perform better than ADDA \cite{ADDA} by a large margin. In particular, embedding both domain-invariance and feature augmentation leads to an improvement $>10\%$. We did not include the work by Haeusser et al. \cite{ADA} in Table~\ref{tab:res} because it makes use of a much more powerful feature extractor (conv-conv-pool-conv-conv-pool-conv-conv-pool-fc-softmax), which makes their method hard to compare with other works.

Finally, Table~\ref{tab:res_di} shows the difference of performance on classifying source samples using $C \circ E_s$ or $C \circ E_I$. As it can be observed, the encoder $E_I$ (trained following Step 2) works well on source samples, too. This allows to use the same encoder for both target and source data, a very useful feature in an application setting where we might not know the source of the data. The worst results on source samples, achieved on SVHN dataset, are most likely due to the large difference between the source and the target domains.     
 
\subsection{Limitations}

The main limit of the domain-invariant feature extractor we designed is the same that can be detected in the works by Tzeng et al. \cite{ADDA} and by Ganin and Lempitsky \cite{Ganin}. Practically, all these approaches encourage source and target features to be indistinguishable, but this does not guarantee that target samples will be mapped in the correct regions of the feature space. In our case and in ADDA's one, this strongly depends on the feature extractor trained on source samples: if the representation is far from being good, the results will be sub-optimal.

\section{Conclusions and future work}\label{sec:conclusion}

In this work, we proposed two techniques to improve the current usage of GAN objectives in the unsupervised domain adaptation framework. First, we induced domain-invariance through a straightforward extension of the original algorithm. Second, we proposed to perform data augmentation in the feature space through GANs \cite{GAN}, a novel application. An exhaustive evaluation was carried out on standard domain adaptation benchmarks, and results confirmed that both approaches lead to higher accuracies on target data. Also, we showed that the obtained feature extractors can be used on source data, too.

Results showed that our approach is comparable or superior to current state-of-the-art methods, with the exception of a single benchmark. In particular, we performed better than recent, more complex methods that rely on generating target images to tackle unsupervised domain adaptation tasks. 
This achievement re-opens the debate on the necessity of generating images belonging to the target distribution: recent results \cite{GOOGLE,UITITN} seemed to suggest it.

For future work, we plan to test our approach on more complex unsupervised domain adaptation problems, as well as investigate if feature augmentation can be applied to different frameworks, \eg, the contexts where traditional data augmentation proved to be successful. 

\section*{Acknowledgments}

The authors would like to thank Lyne P. Tchapmi for helpful suggestions on an early draft. The research reported in this publication was supported by funding from MURI (1186514-1-TBCJE). 

\clearpage

{\small
\bibliographystyle{ieee}
\bibliography{egbib}
}

\onecolumn
\appendix
\section{Architectures}
\begin{figure}[b!]
	\begin{center}
		\includegraphics[width=0.8\textwidth]{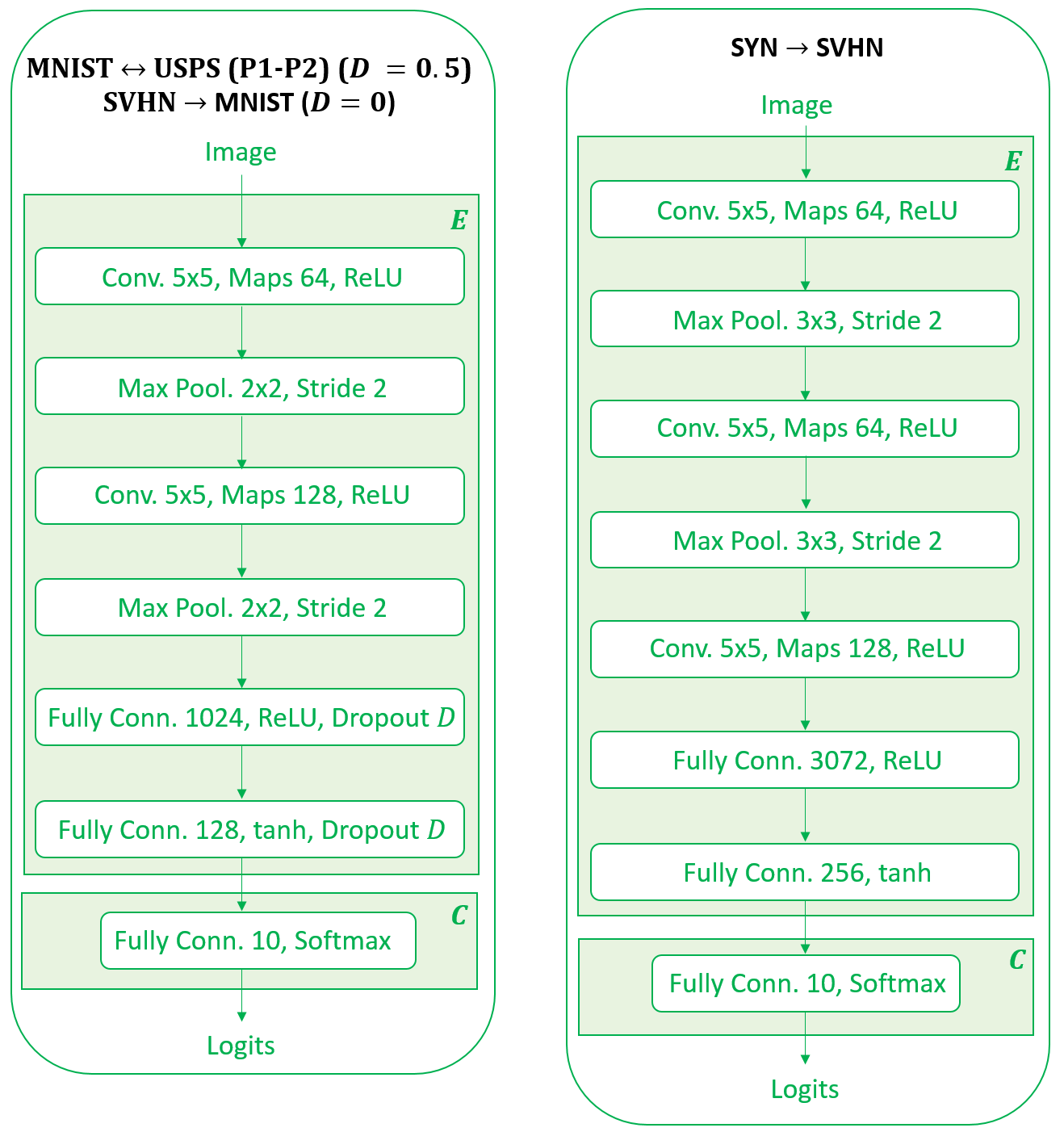}
	\end{center}
	\caption{Architectures used for $C\circ E_S$ and $C\circ E_I$ ($C\circ E$ for simplicity) in the MNIST $\leftrightarrow$ USPS (P1-P2) and in the SVHN $\rightarrow$ MNIST (\textit{left}) experiments, with the different values of Dropout \cite{Dropout} indicated ($D$), and in the SYN $\rightarrow$ SVHN experiment (\textit{right}). The classification module ($C$) is a simple fully-connected + softmax layer.} 
	\label{fig_1}
\end{figure}

\begin{figure}[t!]
	\begin{center}
		\includegraphics[width=0.85\textwidth]{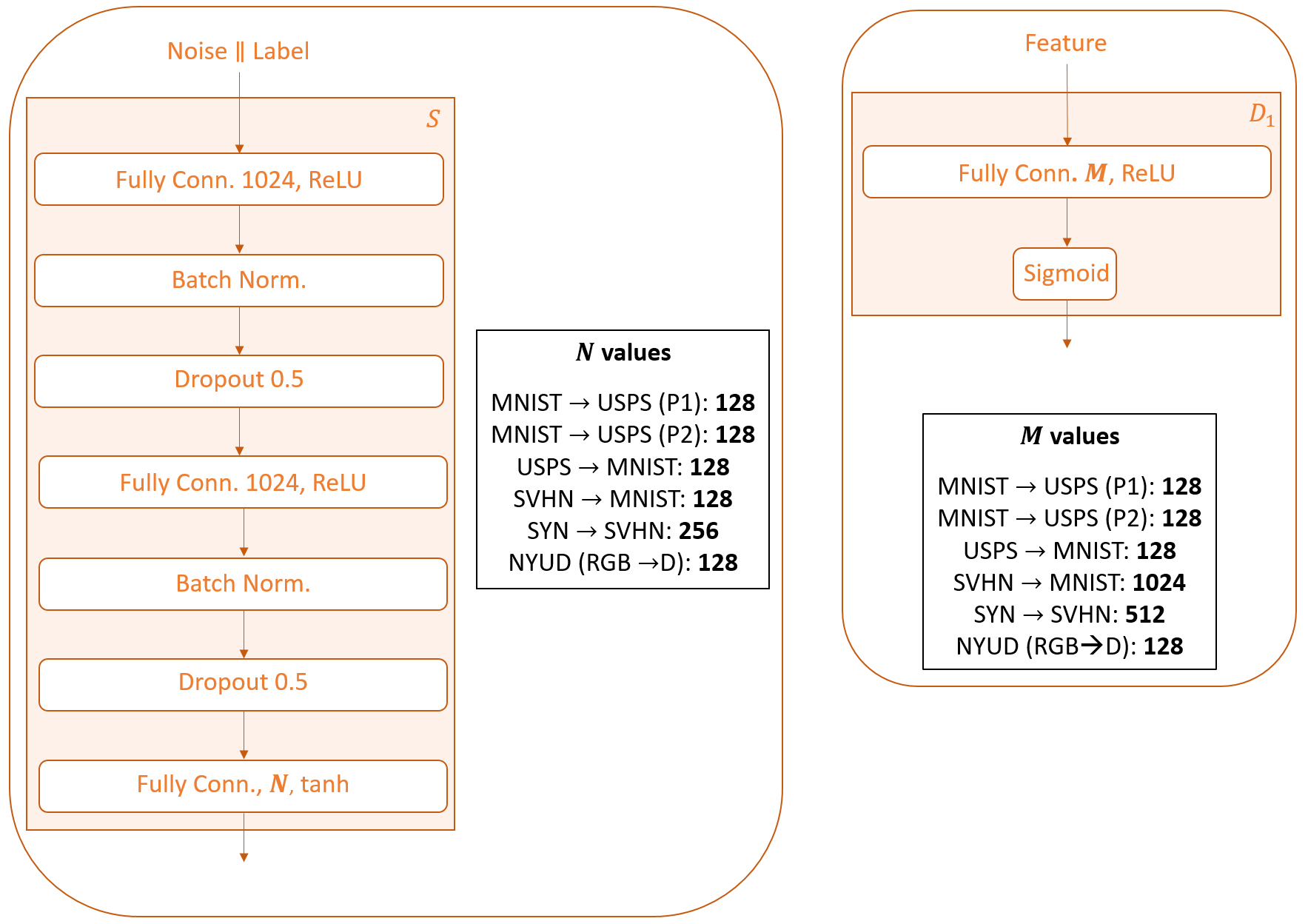}
	\end{center}
	\caption{Architectures used for $S$ (\textit{left}) and for $D_1$ (\textit{right}), with the size of the features generated and of the hidden layer indicated, respectively.} 
	\label{fig_2}
\end{figure}

\begin{figure}[t!]
	\begin{center}
		\includegraphics[width=0.7\textwidth]{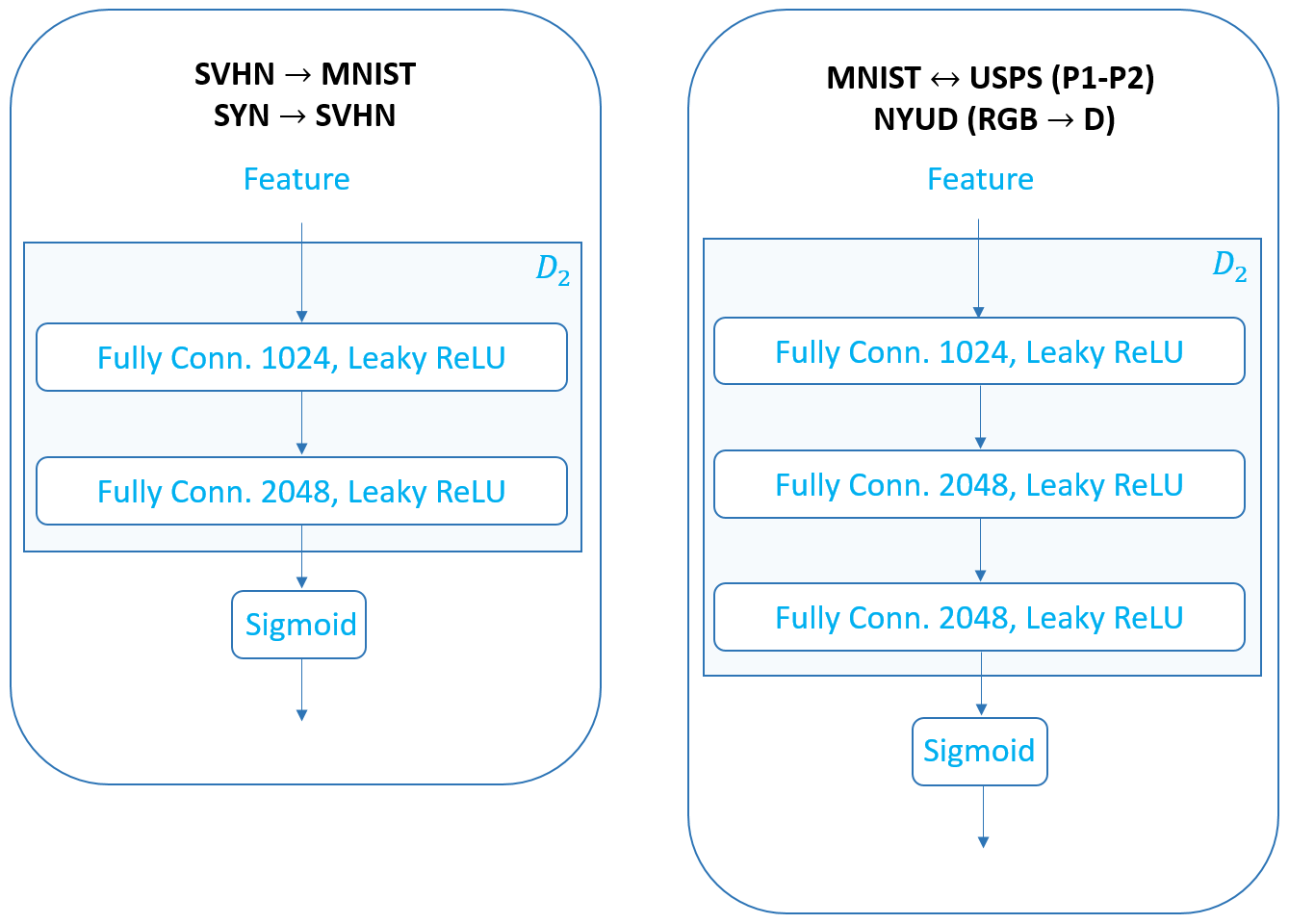}
	\end{center}
	\caption{Architectures used for $D_2$ in the NYUD and MNIST $\leftrightarrow$ USPS experiments (\textit{right}) and in all the others (\textit{left}).} 
	\label{fig_3}
\end{figure}

We provide in this section a detailed description of the networks used for our experiments. For the digit datasets, the encoders follow the standard architectures commonly used in unsupervised domain adaptation \cite{Ganin}.\\~\\
Figure~\ref{fig_1}, \textit{left}: architectures of $E_S$ and $E_I$ used for MNIST $\leftrightarrow$ USPS and SVHN $\rightarrow$ MNIST.\\~\\
Figure~\ref{fig_1}, \textit{right}: architectures of $E_S$ and $E_I$ used for SYN $\rightarrow$ SVHN.\\~\\
Figure~\ref{fig_2}, \textit{left}: architecture of $S$ used for all the experiments.\\~\\
Figure~\ref{fig_2}, \textit{right}: architecture of $D_1$ used for all the experiments.\\~\\
Figure~\ref{fig_3}, \textit{left}: architecture of $D_2$ used for SVHN $\rightarrow$ MNIST and SYN $\rightarrow$ SVHN.\\~\\
Figure~\ref{fig_3}, \textit{right}: architecture of $D_2$ used for MNIST $\leftrightarrow$ USPS and NYUD (RGB $\rightarrow$ D).\\~\\
Concerning $E_S$ and $E_I$ used in the NYUD experiment, we relied on a pretrained VGG-16 \cite{VGG16}, following the protocol used by Tzeng et al. \cite{ADDA}. We cut it at \textit{fc7}, which was shrieked to be 128-dim and modified with tanh activations. The classifier $C$ consists in an additional 19-dimensional softmax layer.\\ 
We found out that $D_2$ should be built with two or three hidden layers to stabilize the minimax game against $E_I$ (whose structure must be the same as $E_S$). We designed an $S$ that proved to be reliable in all experiments; to play a balanced minimax game, we found out that a one-hidden-layer neural network as a discriminator ($D_1$) is an optimal choice. The size of the hidden layer depends on the problem, and can be determined by observing the stability of the training procedure. 

\section{Hyperparameters}
We report in this section the hyperparameters used in the different Steps of the training procedures. Note that hyperparameters were set in order to reach the convergence of the GAN \cite{GAN} minimax games, no cross-validation using target labels was performed. 
\subsection{Digits}
For each training Step, we used a batch size of $64$ samples. The learning rate was set to $3 \cdot 10^{-4}$ for Step 0, $1 \cdot 10^{-4}$ for Step 1 and $3 \cdot 10^{-5}$ for Step 2, in all experiments except MNIST $\leftrightarrow$ USPS, where was set to $3 \cdot 10^{-6}$.

\subsection{NYUD}
In Step 0, the network is not trained from scratches: following the protocol described in \cite{ADDA}, we fully fine-tune a VGG-16 network \cite{VGG16} (pre-trained on ImageNet \cite{ImageNet}) for 20.000 iterations, in order to have a comparable baseline model. Batch size is 32 (instead  of 128) due to hardware limitations. The learning rate were $10^{-4}$ for Step 0, $10^{-5}$ for Step 1 and $10^{-7}$ for Step 2.

\end{document}